**Dynamic Benchmarks: Spatial and Temporal Alignment for ADS Performance Evaluation**

**Yin-Hsiu Chen**
Data Scientist
Waymo LLC
Mountain View, CA, 94043
Email: yinhsiuchen@waymo.com

**John M. Scanlon**
Safety Researcher
Waymo LLC
Mountain View, CA, 94043
Email: johnscanlon@waymo.com

**Kristofer D. Kusano**
Safety Researcher
Waymo LLC
Mountain View, CA, 94043
Email: kriskusano@waymo.com

**Timothy L. McMurry**
Data Scientist
Waymo LLC
Mountain View, CA, 94043
Email: tmcmurry@waymo.com

**Trent Victor**
Director of Safety Research and Best Practices
Waymo LLC
Mountain View, CA, 94043
Email: trentvictor@waymo.com

Word Count: 4818 words + 2000 words (3 table + 5 figures, 250 words per table and figure) = 6818 words

*Submitted on 2024-07-31*

**ABSTRACT**
Deployed SAE level 4+ Automated Driving Systems (ADS) without a human driver are currently operational ride-hailing fleets on surface streets in the United States. This current use case and future applications of this technology will determine where and when the fleets operate, potentially resulting in a divergence from the distribution of driving of some human benchmark population within a given locality. Existing benchmarks for evaluating ADS performance have only done county-level geographical matching of the ADS and benchmark driving exposure in crash rates. This study presents a novel methodology for constructing dynamic human benchmarks that adjust for spatial and temporal variations in driving distribution between an ADS and the overall human driven fleet. Dynamic benchmarks were generated using human police-reported crash data, human vehicle miles traveled (VMT) data, and over 20 million miles of Waymo's rider-only (RO) operational data accumulated across three US counties. The spatial adjustment revealed significant differences across various severity levels in adjusted crash rates compared to unadjusted benchmarks with these differences ranging from 10% to 47% higher in San Francisco, 12% to 20% higher in Maricopa, and 7% lower to 34% higher in Los Angeles counties. The time-of-day adjustment in San Francisco, limited to this region due to data availability, resulted in adjusted crash rates 2% lower to 16% higher than unadjusted rates, depending on severity level. The findings underscore the importance of adjusting for spatial and temporal confounders in benchmarking analysis, which ultimately contributes to a more equitable benchmark for ADS performance evaluations.

**INTRODUCTION**

Benchmarking SAE level 4+ Automated Driving Systems (ADS) is a growing area of research, which compares the historical, operational performance (without a human behind the wheel) of ADS technology to some benchmark (1, 2, 3, 4, 5, 6, 7, 8). This type of retrospective safety impact assessment is part of a safety determination lifecycle, serving to increase the continuous confidence growth of the safety determination process as a whole (9).

There are a number of key technical challenges when benchmarking ADS performance (10), including accounting for reporting practices and exposure differences in Operational Design Domain (ODD) driving. At its core, a safety impact assessment needs to compare a crash population, normalized by some exposure (usually driving distance), between two populations being compared. In order to enhance comparability of crash rates between two populations, both the numerators (crash populations) and denominators (exposure) should be aligned as closely as possible. As part of this alignment, the comparison should strive to account for influential features in the data that may bias the results and potentially lead to an incorrect conclusion (10). The ODD is one key feature to align and can include elements such as vehicle body type, roadway type, weather conditions, traffic density, temporal factors (e.g. time of day and seasonality), and geographic location. Human crash data typically report many of these dimensions, whereas human vehicle miles traveled (VMT) data have less rich reporting, which restricts the ability to perform a precise alignment.

Recent ADS benchmarking has attempted to subset the benchmark crash and VMT data to match the ADS ODD. Scanlon et al. (7) developed benchmarks from human data to match an ADS ride-hailing ODD in San Francisco, Los Angeles, and Phoenix. This benchmark used police-reported state and national crash data and state and federal VMT data to develop a crash rate benchmark in geographic areas (counties) where ADS ride-hailing services currently operate (on non-freeway roads) and involved passenger vehicles (the same as currently deployed ADS ride-hailing platforms). Flannagan et al. (3) used insurance and telematics data from a rental passenger vehicle fleet on surface streets primarily used for ride-hailing to estimate a human benchmark for San Francisco county. Di Lillo et al. (2) compared third party insurance claims rates for Waymo's ride-hailing operations to a human benchmark weighted estimate. The benchmark insurance claims rates were aggregated at the policyholder zip code level and reweighted proportionally by the miles traveled by the Waymo vehicles per zip code. This zip code weighted benchmark provided an approximate adjustment for crash rates dependency on driving location. No road type correction was performed by Di Lillo et al. (2), which provided a conservatively high estimate of the true benchmark crash rate on surface streets.

The current study focuses on the usage of police-reported crash and public driving VMT data in benchmarking ADS performance with spatial and time-of-day alignment. Multiple previous studies have used crash rate performance at a geographic level, such at a county-level, to estimate a crash rate benchmark (7, 11, 12, 13). These prior efforts did not control for variations within that geographic area for the temporal and spatial distribution of driving between both the ADS and benchmark population. Specifically, existing ADS deployments, like Waymo, operate as a paid ride-hailing service mainly on surface streets, and the path these vehicles take are designed to pick-up and move passengers between their needed destinations. This commercial application of the technology dictates where and when the vehicle operates. To that end, past research on human crash data has shown that traffic density (14), time-of-day (15, 16), roadway characteristics (17, 18), and driver demographics (19) can all affect crash rates. The underlying hypothesis of this work is that adjusting for spatial and temporal variations will account for some of the effects of these known, potential confounders. Additionally, as the analysis will demonstrate, the expansion of the fleet and development of the technology has resulted in an evolving ODD over time and, accordingly, changes in the underlying driving risk.

The purpose of this study was to extend the work by Scanlon et al. (7) who did county-level crash rates, and present a methodology for a dynamic human benchmark. The dynamic benchmark adjusts for the spatial and temporal differences between the ADS and benchmark populations. The purpose of this calibration is to determine the benchmark crash rate if the benchmark data had been collected from the same spatial and temporal distribution of driving as the Waymo data. The term "dynamic" refers to the fact that distribution of driving varies over time. So, when evaluating some temporal subset of ADS driving, it is important to match the unique distribution of that driving to ensure an apples-to-apples comparison.

We focused on several research questions. First, how does the geographic and time-of-day driving distribution of the current Waymo ADS fleet compare to the overall driving distribution of the overall human crash population (benchmark) within deployed counties? Second, if the human crash population had driven with the same spatial and temporal distribution as the ADS, what would be the human crash rate (the dynamic benchmark adjustment)? Third, how has service expansion affected this dynamic benchmark adjustment?

## METHODS
### Data Sources
*Human Benchmark Mileage & Crashes*

This analysis relied exclusively on publicly-accessible crash and VMT data to generate crash rate benchmarks. For the crash data, we utilized 2022 police-reported state crash data, specifically focusing on Maricopa (Arizona), Los Angeles (California), and San Francisco (California) counties where Waymo ride-hailing service has been deployed. In Maricopa County, the data source relied upon was the comprehensive state census of all police-reported crashes compiled by the Arizona Department of Transportation (ADOT) accessible through a records request (20, 21, 22). For San Francisco and Los Angeles, we leveraged crashes resulting in injury from the Statewide Integrated Traffic Records System (SWITRS) accessible through the California Highway Patrol's online portal (23). Unlike ADOT crash data, only crashes resulting in injury are required to be reported within SWITRS under California Vehicle Code §20008, so "some agencies report only partial numbers of their PDO crashes, or none at all" (23). Because of this limitation, it is unclear how representative San Francisco and Los Angeles counties are of all police-reported events within those areas, and there is the potential for underreporting in the police-reported crash rate of non-injury human benchmark crashes (7).

Multiple sets of crash data annotations were incorporated into this study's analyses. First, current Waymo rider-only (RO) operations are vastly dominated by surface street driving and hence only crashed passenger vehicles traveling on surface streets were included. The classification routines, including exact variable and value pairings, for vehicle body type and road type using ADOT and SWITRS crash data are described by Scanlon et al. (7). Second, this study examined multiple crash severity levels, which can be found in Table 1. The classification routines for these are also presented in Scanlon et al. (7), and are intended to cover the full spectrum of reportable crash outcomes severity (*police-reported* to *fatal*). Third, to facilitate the calculation of human crash rates at a sub-county level for the derivation of dynamic human benchmarks, we extracted precise geo-locations of individual crashes based on longitude-latitude pairs or road name information. Some events in SWITRS were missing longitude-latitude pairs, which were then determined by entering the crash city ("CNTY_CITY_LOC") and encoded roadways where the crash occurred on ("PRIMARY_RD" and "SECONDARY_RD") into Google's geocoding API.

The United States Department of Transportation (USDOT) Federal Highway Administration (FHWA) compiles the Highway Performance Monitoring System (HPMS) database, mandating comprehensive reporting of road characteristics, including highway functional classification, and vehicle miles traveled (VMT) for all public roads. HPMS employs a stratified sampling methodology to collect traffic volume for road segments, deriving Annual Average Daily Traffic (AADT) estimates. The road-level HPMS data, publicly available for download (24), encompass AADT and geo-locations, with the exception of rural minor collector and local roads (25). In California, missing minor and local roads are rectified by a fuel consumption model, while in Arizona, adjustments are based on historical travel patterns and population growth in non-metropolitan planning organization areas. Additionally, as HPMS data lack road-level AADT specifically for passenger vehicles, we utilize all available mileage data to derive relative VMT across locations by aggregating the product of road length and AADT. We then apply a calibration factor to align this relative VMT with passenger-vehicle-adjusted aggregate mileage filtered by roadway and county criteria. Further details on the extraction of total mileage are reported in Scanlon et al. (7, 26).

**TABLE 1: A list of crash severity levels included in this study that were originally presented in Scanlon et al. (7).**

| Crash Severity Level | Description | Underreporting Adjustment | Inclusion Criteria | |
|---|---|---|---|---|
| | | | Variable | Value |
| *Police-Reported* | A police report was filed (record was present within ADOT and SWITRS). | None | All severity levels. | |
| *Any-Reported-Injury* | Any involved person throughout the crash sequence sustained any level of injury with NHTSA underreporting adjustment applied. | 32% underreporting adjustment (25) | SWITRS: collision_severity | IN ('K', 'A', 'B', 'C') |
| | | | ADOT: InjuryStatus | |
| *Air Bag Deployed* | Any involved vehicle had an airbag deployment. | None | SWITRS: victim_safety_equip_1 OR victim_safety_equip_2 OR party_safety_equip_1 OR party_safety_equip_2 | IN ('L', 'M') |
| | | | ADOT: Airbag | IN (2, 3, 4, 5, 102, 103, 105) |
| *Suspected Serious Injury+* | Any involved person throughout the crash sequence sustained a fatal or suspected serious injury. | None | SWITRS: collision_severity | IN ('K', 'A') |
| | | | ADOT: InjuryStatus | |
| *Any Fatality* | Any involved person throughout the crash sequence sustained a fatal injury. | None | SWITRS: collision_severity | ='K' |
| | | | ADOT: InjuryStatus | |

Time-of-day was the second dynamic benchmark dependency explored. HPMS data does not contain AADT (used to compute VMT) with respect to time, such as by time of day, day of week, or month. This limited our ability to jointly incorporate geographical and temporal factors into the computation of a dynamic benchmark adjustment. To examine the effect of time-of-day, this study incorporated VMT by time period estimates for the year 2016, reported by the San Francisco County Transportation Authority (28), which allowed us to compare temporal variations at discrete time-of-day intervals within San Francisco county. SWITRS crash data from 2016 was also used for this analysis. The aggregate VMTs are derived from lengths of network links, each weighted by the corresponding number of vehicle trips on these links and they do not distinguish different road types. Therefore, we used the VMT distribution across time categories on all road types to approximate the distribution for surface streets. Furthermore, as aggregate VMT does not differentiate between passenger and non-passenger vehicles, we did not exclude crashes involving non-passenger vehicles in time-based analysis. Time-of-day adjustment of dynamic benchmarks in Maricopa county and Los Angeles county was not within the scope of this study due to lack of data.

*Waymo ADS Mileage*

Waymo's autonomous ride-hailing service operates in geographical areas across Phoenix, San Francisco, Los Angeles, and Austin without a human behind the wheel (RO). Waymo's current (as of the end of June 2024) ODD encompasses nearly all non-limited-access roads with speed limits up to 50 mph, with limited restrictions. While several ODD expansions have occurred periodically, the overall ODD for Waymo RO service has remained largely consistent since the end of 2023. The longitudes and latitudes of Waymo vehicle locations are recorded and the precise traveled distance is calibrated based on odometer readings from the rear wheel. Due to limited operational scale in Austin, RO miles from the region are excluded from the analysis. In this study's computation of dynamic human benchmarks, we include an accumulated total of 21.9 million RO miles collected up to 2024-06-30, across the other three areas.

**S2 Cell Framework**

We employed the S2 cell framework (29) to discretize geographical areas into smaller units for derivation of dynamic benchmarks with geo-adjustment. S2 cell is a hierarchical geospatial indexing system developed by Google to partition Earth's surface into a uniform grid of cells.

Theoretically, given precise crash locations and complete road-level driving mileage data for minor arterials and rural roads, a finer S2 cell resolution would be preferable in geo-adjustment to more accurately account for micro-level variations in mileage mix. Acknowledging potential inaccuracies in police-reported crash locations and the absence of granular HPMS data on minor arterial and local roads, we opted for a coarser resolution at S2 cell level 13, characterized by an average area of approximately 1.27 square kilometers, in our analysis. Approximately 0.01% RO miles fall within level-13 S2 cells that lack corresponding lane-level VMT from HPMS. These cells were excluded from our analysis.

Each road segment in the HPMS data may intersect with multiple S2 cells. In such cases, we proportionally allocate VMT of the segment based on the fraction of its length that overlaps with each cell.

**Dynamic Human Benchmarks**

The dynamic benchmark analysis performed in this study focused on spatial (S2 cell) and temporal (time-of-day) adjustments of benchmark crash rates according to the distribution of ADS driving mileage. The dynamic benchmark can be interpreted as representing the *hypothetical human benchmark crash rate if the human benchmark mileage distribution mirrored that of Waymo*. This subsection presents the framework for doing this correction that can be generalized to other factors not examined in this study. To demonstrate this methodology, consider that the driving can be subdivided into slices, as denoted by $s = 1, ..., S$, where each slice contains some aggregation of driving distance ($m_s^{(H)}$) and crash counts ($c_s^{(H)}$) for the human benchmark (H). The human crashed vehicle rate ($\lambda_s^{(H)}$ in incidents per VMT) in equation 1 for a given slice is represented as the ratio of crash counts to driving distance.

$$\lambda_s^{(H)} = \frac{c_s^{(H)}}{m_s^{(H)}} \quad (1)$$

The dynamic benchmark (equation 2) for a given area is computed by taking a weighted average of all slices of driving, where the weighting is proportional to the amount of ADS mileage within the driving slice ($m_s^{(W)}$).

$$Dynamic\ Benchmark = \frac{\sum_{s=1}^{S} m_s^{(W)} \cdot \lambda_s^{(H)}}{\sum_{s=1}^{S} m_s^{(W)}} \quad (2)$$

The dynamic benchmark can also be expressed according to the crash rate contribution of each slice toward the net benchmark rate using total mileage (first term in equation 5 where $M^{(H,W)} := \sum_{s=1}^{S} m_s^{(H,W)}$) and the ratio of proportion of ADS mileage to proportion of VMT driven within that slice (second term in equation 5 where $f_s^{(H,W)} := \frac{m_s^{(H,W)}}{M^{(H,W)}}$):

$$Dynamic\ Benchmark\ =\ \sum_{s=1}^{S} \left( \frac{m_s^{(W)} \cdot c_s^{(H)} / m_s^{(H)}}{M^{(W)}} \right) \qquad (3)$$

$$=\ \sum_{s=1}^{S} \left( \frac{c_s^{(H)}}{M^{(H)}} \right) \cdot \left( \frac{m_s^{(W)} / M^{(W)}}{m_s^{(H)} / M^{(H)}} \right) \qquad (4)$$

$$=\ \sum_{s=1}^{S} \left( \frac{c_s^{(H)}}{M^{(H)}} \right) \cdot \left( \frac{f_s^{(W)}}{f_s^{(H)}} \right) \qquad (5)$$

The expression of dynamic benchmark in equation 5 can be interpreted as a weighted average of slice-level human benchmark contributions, where the weights are proportional to the ratio of Waymo's driving propensity within each slice relative to human VMT propensity in that slice.

**Confidence Intervals**
All confidence intervals presented in this paper are estimated using Poisson bootstrap method (28) with 90% confidence level. For each of the bootstrap iterations (N=1000), a random number from a Poisson distribution with mean 1 ($\lambda = 1$) is generated for each crash event to represent the frequency of that particular event within the resampled data set. The distribution of a quantity of interest (unadjusted benchmark, dynamic benchmark, or dynamic benchmark multiplier) across all bootstrap samples serves as an approximation of the sampling distribution. The 5th and 95th quantiles of the distribution are then used to estimate the lower bound and upper bound, respectively, of the 90% confidence intervals.

**RESULTS**
**Spatial distribution of Benchmark and ADS Driving Data**
Figure 1 presents the spatial distribution of Waymo RO mileage and HPMS VMT across each geographic region. It is notable that the Waymo ADS mileage covers the full extent of San Francisco county, but only select proportions of Los Angeles and Maricopa Counties. A positive correlation is evident, with both Waymo and human drivers traversing more miles in higher-density areas and on arterial roads. However, distinct differences exist in the precise location of hotspots and the dispersion of mileage clustering. Generally, Waymo RO mileage was observed to be more concentrated in commercial areas and routes connecting compact urban areas, compared to the more evenly distributed HPMS VMT, which included a higher proportion of mileage in residential areas.

| Geo | Waymo RO Mileage | HPMS VMT |
| --- | --- | --- |

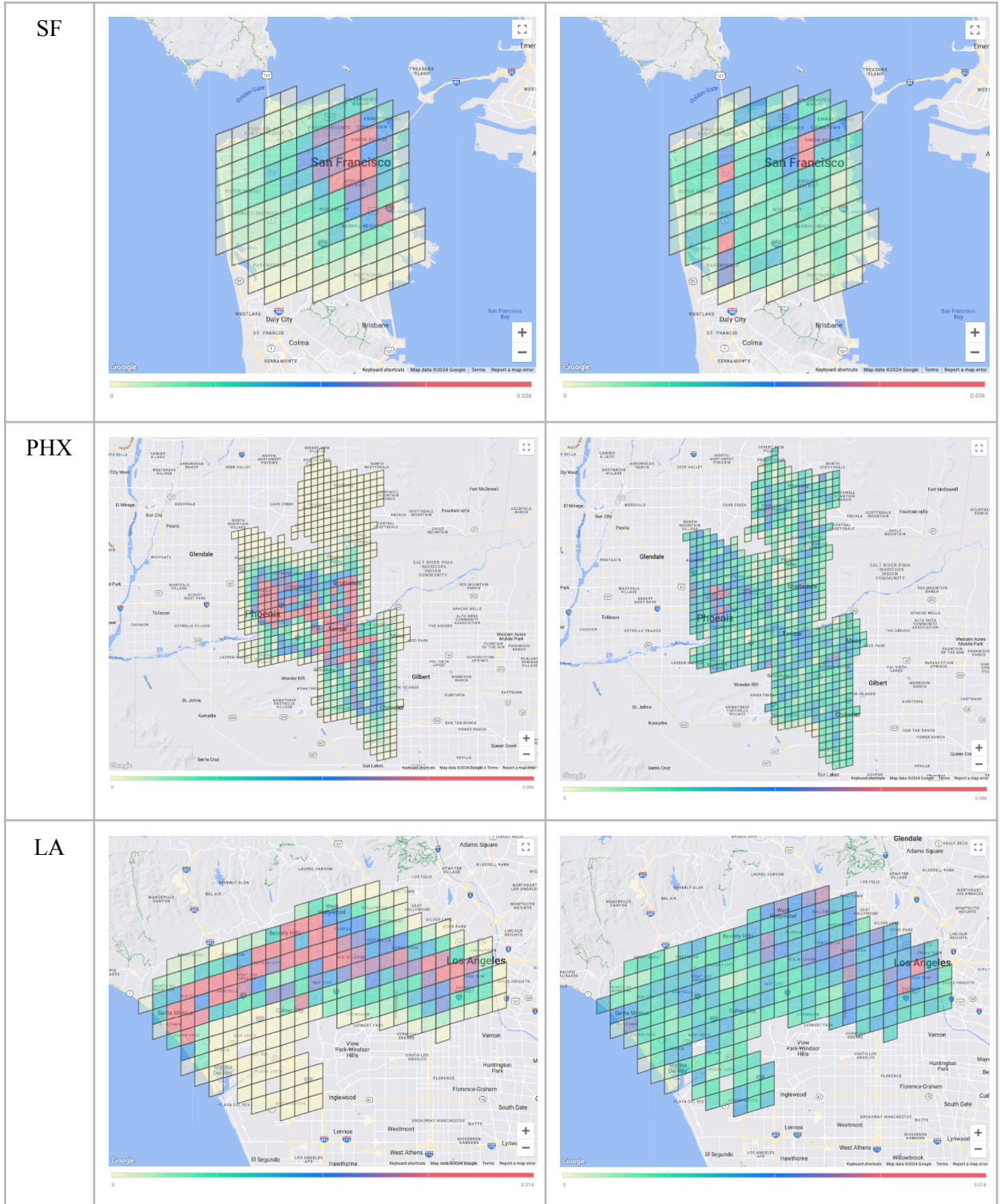

**Figure 1: Heatmaps illustrate Waymo RO mileage (left column) and HPMS VMT (right column). Each geographic region's depicted dual heatmaps share the same scale with individual cells representing the proportional mileage within that cell relative to the entire geo. Light colors correspond to less driving, with blue representing middle range, while red corresponds to more driving.**

**Crash Rate Dependencies on Spatial Distributions**

Figure 2 depicts the distribution of Waymo mileage versus RO miles across S2 cells categorized by police-reported crash rate, denoted by 10 quantile buckets. Quantile buckets with a low value have lower unadjusted crash rate and buckets with higher value have a higher unadjusted crash rate. Waymo RO mileage in San Francisco and Phoenix is disproportionately concentrated in buckets with higher human crash rates. While the trend is less pronounced in Phoenix, Los Angeles only exhibits a higher relative portion of Waymo mileage in the top bucket, with a mixed distribution in the remaining categories. This result demonstrates the interaction between the differences in driving distributions and the correlation with benchmark crash rates. Consequently, a direct comparison of ADS crash performance against a human benchmark without adjusting for this differential exposure could underestimate (bias) the potential risks encountered by ADS in San Francisco and Phoenix.

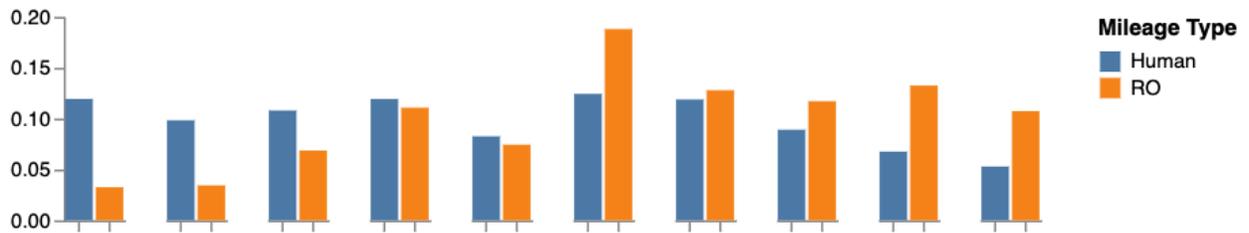

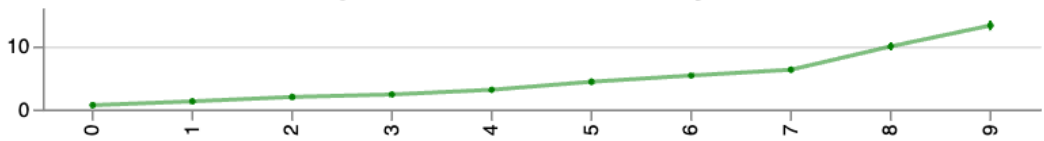

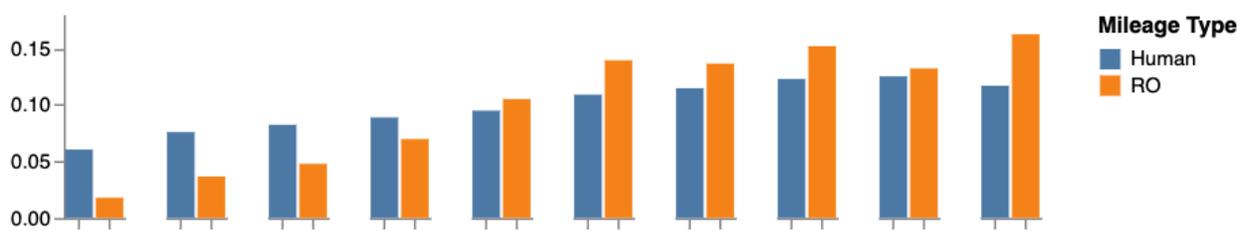

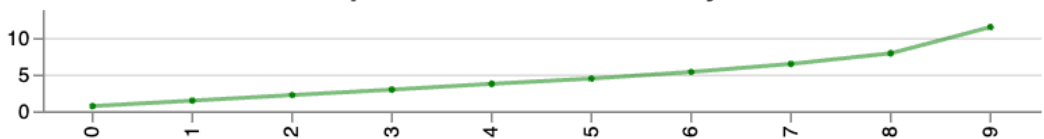

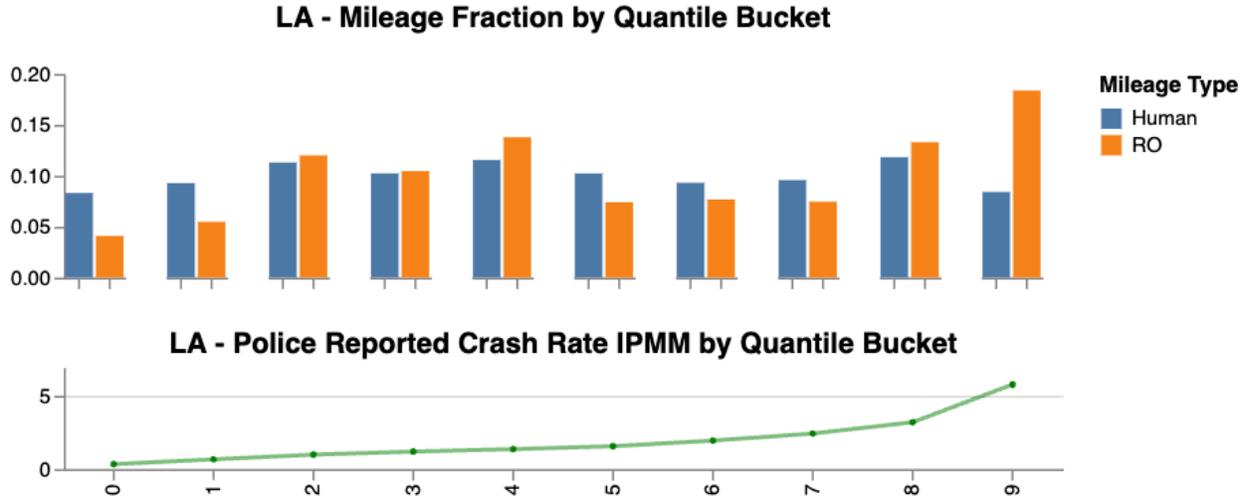

**Figure 2: Distribution of human and RO mileage fractions across quantile buckets of *police-reported* crash rates in San Francisco (upper), Phoenix (middle), and Los Angeles (lower).**

Figure 3 presents a comparison of dynamic human benchmarks (with spatial-adjustment) with their unadjusted counterparts (without spatial-adjustment) for each severity level evaluated. In San Francisco and Maricopa counties, dynamic benchmarks were significantly higher than unadjusted human benchmarks across all severity levels except for fatal crashes, which had a limited, low sample size of crash events. In Los Angeles County, the dynamic benchmark *police-reported*, *any-injury-reported*, and *airbag deployed* crash rates were higher than the unadjusted crash rates. There were no differences between unadjusted and dynamic benchmarks for *suspected serious injury+* and *fatal* crashes in Los Angeles.

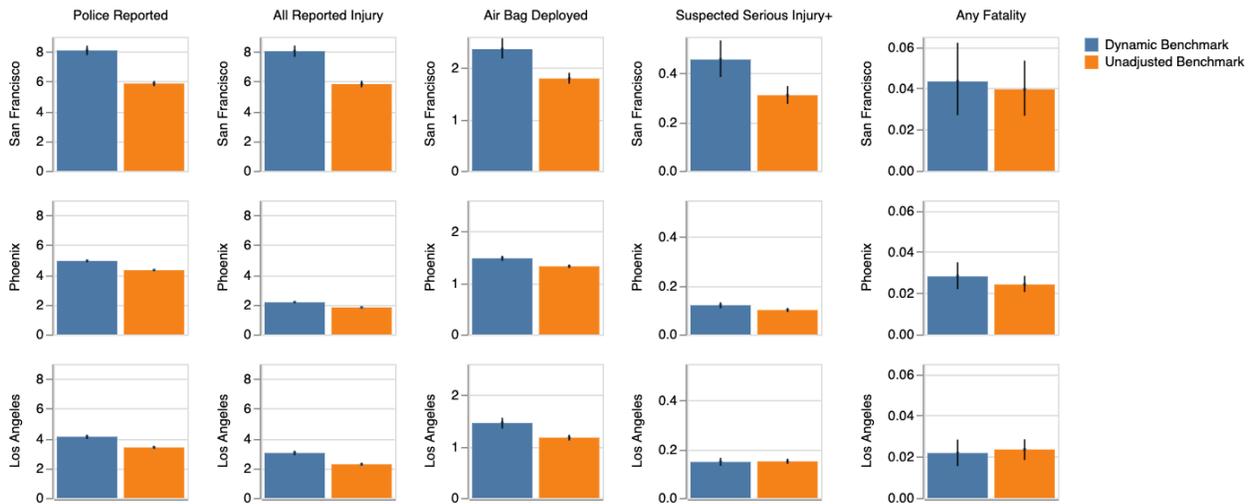

**Figure 3: Comparison of human benchmark crash rates per million miles for the dynamic (blue) and unadjustment (orange) benchmarks for various crash outcome groups in San Francisco, Phoenix, and Los Angeles.**

The impact of the adjustment is more pronounced in San Francisco than Maricopa county and Los Angeles. This is further quantified in Table 2, which presents dynamic benchmark correction multipliers, defined as the ratio of the dynamic benchmark to the unadjusted benchmark. For instance, the unadjusted police-reported crash rates in San Francisco and Maricopa county are 5.86 (90% CI: [5.68, 6.03]) IPMM and 4.31 (90% CI: [4.26, 4.36]) IPMM, respectively. After applying reweighting based on Waymo RO miles, these crash rates increase to 8.08 (90% CI: [7.77, 8.40]) IPMM and 4.91 (90% CI: [4.84, 5.00]) IPMM, respectively, revealing a human benchmark underestimation of 1.38 and 1.14 times when not performing a spatial dynamic benchmark correction.

**TABLE 2: dynamic benchmark correction multipliers (90% confidence intervals), defined as the ratio of dynamic human benchmark and unadjusted human benchmark, for various crash outcomes in San Francisco, Phoenix, and Los Angeles.**

|  | San Francisco | Phoenix | Los Angeles |
|---|---|---|---|
| **Police Reported** | 1.38 [1.35, 1.41] | 1.14 [1.13, 1.16] | 1.21 [1.18, 1.24] |
| **Any Injury Reported** | 1.38 [1.34, 1.42] | 1.19 [1.16, 1.22] | 1.34 [1.30, 1.37] |
| **Air Bag Deployed** | 1.32 [1.25, 1.40] | 1.12 [1.09, 1.14] | 1.24 [1.18, 1.30] |
| **Suspected Serious Injury+** | 1.47 [1.32, 1.65] | 1.20 [1.11, 1.28] | 0.99 [0.91, 1.07] |
| **Any Fatality** | 1.10 [0.88, 1.34] | 1.16 [1.00, 1.34] | 0.93 [0.72, 1.15] |

**Effect of Service Expansion on Dynamic Human Benchmarks**

Waymo's expansion of RO service territories has followed a phased, gradual expansion across various regions. In San Francisco, the first 100,000 operational miles were primarily concentrated in the Sunset neighborhood of San Francisco in the southwest of the city. The area has lower average human crash rates compared to other regions in the city, where Waymo subsequently expanded its service, with the northeastern region being the most recent addition (Figure 4). This area has become a high-demand zone for Waymo's RO service and represents a large proportion of the recent (observable in January and June 2024) overall mileage accumulation. Analysis of the latter half of cumulative mileage in San Francisco suggests a stabilization of the spatial distribution of mileage.

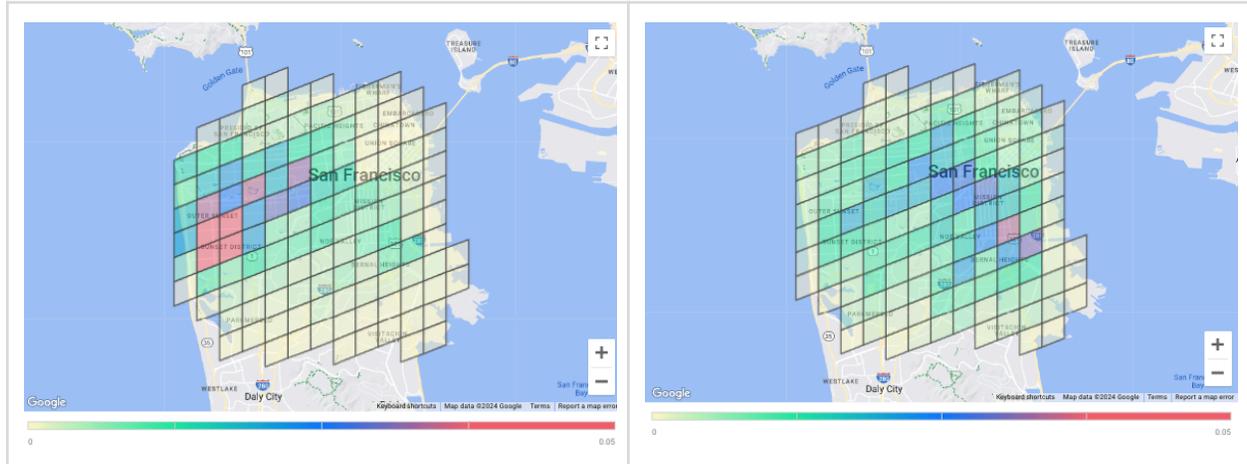

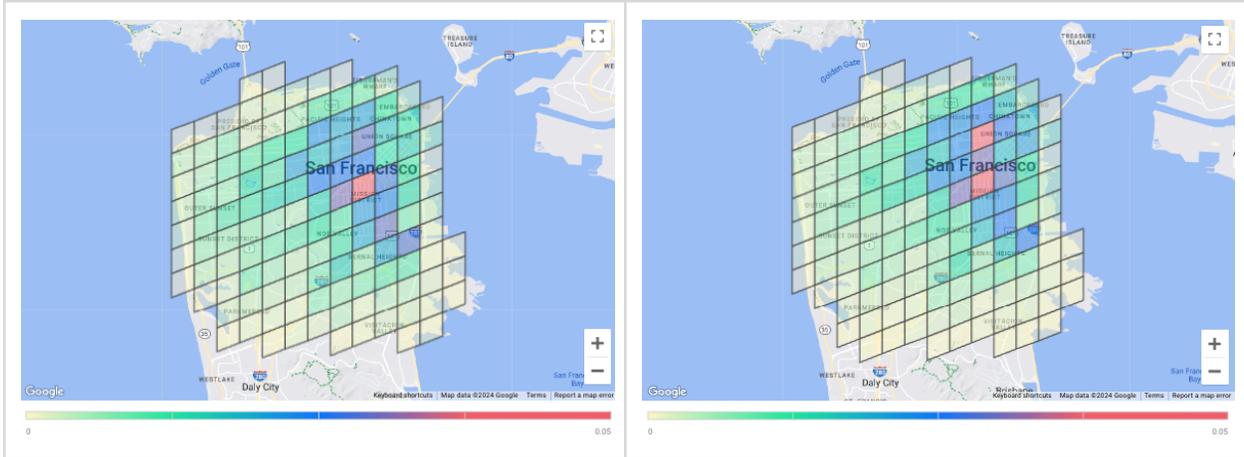

**Figure 4: Heatmaps illustrating the evolution of Waymo RO mileage in San Francisco across four milestones: 100,000 miles (January 2023) in upper left, 1 million miles (July 2023) in upper right, 2.9 million miles (January 2024) in lower left, and 5.9 million miles (June 2024) in lower right. Each cell represents proportional mileage within that cell to the entire city and the quadruplet of the maps are on the same scale.**

The progressive expansion of Waymo's RO service in San Francisco has led to a notable shift on dynamic benchmark correction multipliers, as shown in Figure 5. Initially below 1.0, these multipliers have increased to values above 1.0 and reached a stable state as the operational territory expanded. This upward shift signifies a change in Waymo's mileage distribution, with a greater proportion of miles driven in areas characterized by higher human crash rates. This trend is consistent across various crash outcome categories.

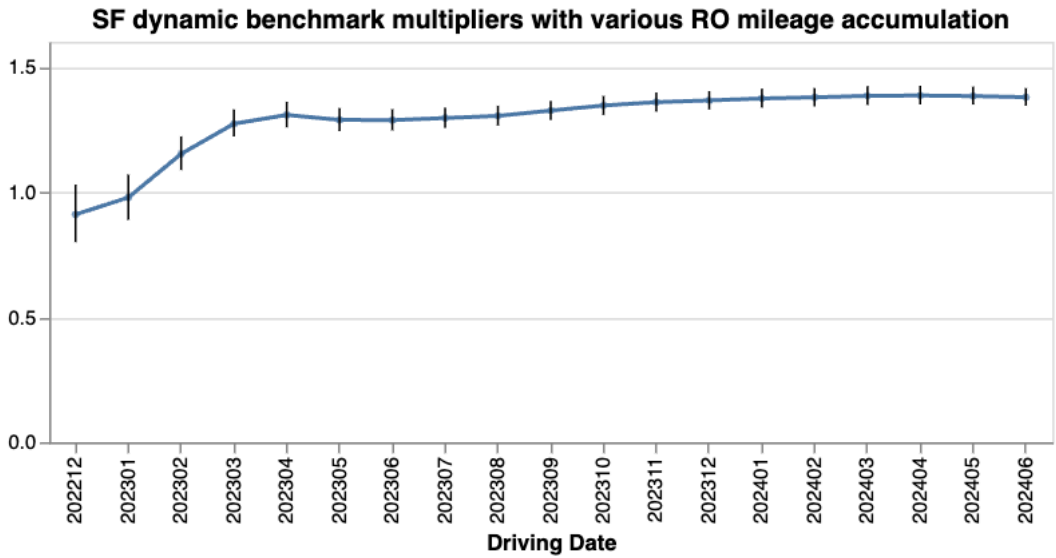

**Figure 5: Time-series of dynamic benchmark correction multipliers based on police-reported crashes in San Francisco as RO mileage accumulates.**

Waymo's RO service in Arizona started with a fleet of Chrysler Pacifica vehicles, which accumulated 909 thousand miles before the Pacifica platform was retired in May 2023. The service area for the Pacifica fleet was restricted to Chandler, Arizona, whereas the current fleet of Jaguar I-PACE vehicles covers a broader area of the greater Phoenix area including downtown and Scottsdale. The discrepancy in the exposure between the Pacifica and I-Pace fleets due to geographic location is noticeable. The dynamic benchmark multiplier, calculated based on police reported crashes, was 1.14 (90% CI: [1.13, 1.16]) for the I-PACE fleet and was 1.01 (90% CI: [0.99, 1.03]) for the Pacifica fleet. Across all other crash outcomes, the I-PACE fleet was associated with higher dynamic human benchmarks than the Pacifica fleet, indicating a lower baseline safety risk in Chandler, where the Pacifica fleet was exclusively deployed. While dynamic benchmark adjustment has a negligible impact on Pacifica fleet's crash rates across all outcomes, the multipliers were statistically significantly greater than 1.0 except fatal crashes for the I-PACE fleet.

**Crash Rate Dependencies on Time-of-day Distributions**

Comparing 2016 daily VMT from Figure 20 in the TNC & Congestion report (28) with RO mileage distribution presented in Table 3 reveals key differences in the time-of-day distribution of driving between the Waymo and human benchmark fleet. Waymo VMT was more concentrated in the late afternoon to early morning, while human benchmark mileage is proportionally higher in the morning through mid-afternoon. Furthermore, higher normalized crash rates, except for fatal crashes, coincide with increased RO mileage during evening and overnight hours, while lower crash rates align with reduced RO mileage during morning commute hours, across all severity levels. Together these considerations suggest that the Waymo ADS operates more at times of elevated non-fatal risk than the average human driver. Applying the dynamic benchmark methodology based solely on time-of-day adjustment yields a crash rate of 7%, 5%, 16%, and 12%, higher than the unadjusted crash rate for police reported crashes, any reported injury, crashes involving air bag deployed, suspected serious injury+, respectively. The adjusted crash rate for fatality is 2% lower than unadjusted crash rate, albeit with a large margin of uncertainty.

**TABLE 3: Time-of-day distribution of Waymo RO miles, human VMT, and relative crash rates for various severity levels in San Francisco.**

| Time Window | Waymo RO Miles (%) | Human VMT in 2016 (%) | 2016 Relative Crash Rate (SWITRS) | | | | |
| --- | --- | --- | --- | --- | --- | --- | --- |
| | | | Police Reported | Any Reported Injury | Airbag Deployed | Suspected Serious Injury+ | Any Fatality |
| Early Morning 3AM - 6AM | 5.6% | 2.8% | 1.31 | 1.16 | 1.87 | 1.07 | 0.92 |
| Morning Commute 6AM - 9AM | 9.0% | 14.9% | 0.66 | 0.68 | 0.52 | 0.54 | 0.34 |
| Late Morning & Early Afternoon 9AM - 3:30PM | 15.7% | 37.9% | 0.96 | 0.99 | 0.87 | 0.74 | 1.15 |
| Late Afternoon 3:30PM - 6:30PM | 29.2% | 20.4% | 0.99 | 1.04 | 0.85 | 1.24 | 1.39 |
| Evening & Overnight 6:30PM - 3AM | 40.5% | 24.1% | 1.25 | 1.16 | 1.52 | 1.48 | 0.85 |
| Dynamic Benchmark Multiplier (90% CI) | | | 1.07 [1.06, 1.08] | 1.05 [1.03, 1.06] | 1.16 [1.13, 1.18] | 1.12 [1.07, 1.17] | 0.98 [0.86, 1.12] |

**DISCUSSION**

Dynamic human benchmark adjustments serve as one potentially valuable tool for facilitating equitable comparisons between ADS and benchmark driving performance. When determining whether an adjustment is necessary, it is important to consider (a) which factors might influence crash rates for either the ADS or benchmark? and (b) how different is the exposure to that influential factor between the ADS and benchmark? This study demonstrated differences in spatial and temporal characteristics of the ADS and general human driving population across three deployment regions. These disparities likely reflect both the staged deployment of Waymo RO services and differences in demand patterns between Waymo's transportation-as-a-service (TaaS) and general human driving. The dynamic benchmark spatial adjustment generally resulted in higher adjusted benchmarks, ranging from 10% to 47% higher in San Francisco county, 12% to 20% higher in Maricopa county, and 7% lower to 34% higher in Los Angeles county, across severity levels. The time-of-day variations were evaluated for San Francisco county only (due to limited human data availability), and the differences were found to result in 2% lower to 16% higher adjusted crash rates, which represents a more modest effect compared to the spatial adjustment.

The limited data availability constrains the ability to adjust for all possible factors that may influence crash rates. Nonetheless, our study demonstrates that spatial and temporal differences are two influential factors affecting collision risk in the counties being examined. The expectation from this adjustment is that this would effectively capture known confounders for crash risk, such as traffic density, time-of-day, roadway characteristics, and driver demographics (14, 15, 16, 17, 18, 19). Future work, data permitting, may consider more explicitly evaluating the effect of other features. Another known limitation of this current study was that, due to data limitations, temporal and spatial effects were independently evaluated. Future work may look at the interaction effect of these two dimensions on crash rates, as there is likely some interaction between geographic and temporal distributions.

Another notable limitation of slice-level calibration for dynamic benchmark derivation is the potential presence of unmeasured confounders within slices. The temporal and spatial adjustments were fairly coarse in nature due to the lack of granular benchmark data available. For instance, the lack of representation of local roads in road-level HPMS data hinders our ability to apply geo-adjustment at a more fine-grained level. Furthermore, there is some inherent imprecision in crash location geocoding in police-reported data that necessitates the selection of a reasonably coarse adjustment level to mitigate potential bias. Generally speaking, coarser S2 cell levels and time-of-day buckets might bring the dynamic benchmark closer to the unadjusted benchmark, thereby risking "washing out" the effect. Future research with access to more granular and quality data could further refine the alignment of ADS mileage and human VMT distributions, thereby enabling a more accurate assessment of dynamic benchmarks.

This study considered spatial and temporal effects, and also performed subselection by road type (surface streets) and vehicle type (passenger vehicles). Additional adjustments can further improve the alignment of the benchmark and ADS driving exposure. Some considerations include weather (31, 32), day-of-week (33), and seasonality (34), which are all known confounders of human driven crash risk and potentially interact with the spatial and temporal effects examined in this study. While Waymo RO services operate in various weather conditions, including low visibility caused by fog and rain, the system does not operate in heavy fog, rain, or blowing sand above certain thresholds. The current ODDs, however, have relatively low exposure rates to such events that cause operating restrictions. Furthermore, there is no human exposure data to do a dynamic benchmark adjustment along this severe weather dimension. Limited driving exposure, data availability, and mixed results from previous studies on the effect of meteorological variables on crash frequency and severity (31, 32) make it challenging to determine how weather conditions might affect dynamic benchmark adjusted crash rates.

The presentation of a dynamic benchmark represents an incremental improvement to the state-of-the-art that, to the knowledge of the authors, had not been executed in benchmarks previously. As the methodology continues to improve, we recommend researchers approach benchmarks cautiously - prioritizing accuracy and erring on the side of conservatism in any estimates (10). Future work should explore novel data sets and methodology to integrate additional dimensions to provide more insights about relative crash risk. Researchers should also continue to examine existing methodology for spatial and temporal corrections, along with ongoing assessment of the effect of influential factors on crash type composition and severity level. Ultimately, dynamic benchmarks can be further enhanced by gaining a deeper understanding of the factors influencing benchmark and ADS crash risk and by employing meaningful, interpretable, and reliable choices when controlling for confounding factors in a dynamic benchmark.

# CONCLUSIONS

Our study compared dynamic benchmarks, adjusted for spatial and hour-of-day factors, to unadjusted county-level human benchmarks. The increased spatial precision in the dynamic benchmarks revealed significant differences in crash rates across various severity levels, indicating that Waymo RO vehicles disproportionately operate in areas with higher human crash risk in San Francisco and Maricopa counties with mixed results observed in Los Angeles County. Time-of-day adjustments computed in San Francisco also resulted in increased non-fatal crash rates, suggesting that Waymo operates more during times of elevated risk in this region. The effect of the spatial adjustment was larger in magnitude in San Francisco than the time-of-day adjustment. Building upon the work of Scanlon et al. (7), this study represents a meaningful step towards addressing the challenge of accounting for Operational Design Domain (ODD) exposure differences in ADS benchmarking. Our findings underscore the importance of utilizing dynamic benchmarks to ensure equitable and accurate comparisons of ADS safety performance to relevant benchmarks. Human data availability limitations restrict the ability to do further adjustments for influential factors. Future research and new data sources should prioritize addressing these limitations by exploring novel data sources and methodologies to enable a more comprehensive understanding of the complex interplay between various factors influencing crash risk and more accurate safety assessments of ADS.

# AUTHOR CONTRIBUTIONS: CREDIT

Yin-Hsiu Chen: conceptualization, data curation, methodology, software, writing - original draft. John M. Scanlon: conceptualization, data curation, software, writing - original draft, writing - review and editing. Kristofer D. Kusano - supervision, writing - original draft, writing - review and editing. Timothy L. McMurry: data curation, writing - review and editing. Trent Victor: supervision, writing - review and editing